\documentclass[final,5p,twocolumn]{elsarticle}

\usepackage{amssymb}
\usepackage{amsthm}
\usepackage{amsmath}
\usepackage{subfigure}
\usepackage{multirow}

\begin{document}

\begin{frontmatter}



\title{Uncertainty-aware deep learning for digital twin-driven monitoring: Application to fault detection in power lines}


\affiliation[inst1]{organization={Reliability and Risk Engineering Lab, Department of Mechanical and Process Engineering},
            addressline={ETH Zurich}, 
            city={Zurich},
            postcode={8092}, 
            country={Switzerland}}

\author[inst1]{Laya Das}

\author[inst1]{Blazhe Gjorgiev}
\author[inst1]{Giovanni Sansavini\corref{*}}
\cortext[*]{Corresponding author, email: sansavig@ethz.ch}


\begin{abstract}
Deep neural networks (DNNs) are often coupled with physics-based models or data-driven surrogate models to perform fault detection and health monitoring of systems in the low data regime. These models serve as digital twins to generate large quantities of data to train DNNs which would otherwise be difficult to obtain from the real-life system. However, such models can exhibit parametric uncertainty that propagates to the generated data. In addition, DNNs exhibit uncertainty in the parameters learnt during training. In such a scenario, the performance of the DNN model will be influenced by the uncertainty in the physics-based model as well as the parameters of the DNN. In this article, we quantify the impact of both these sources of uncertainty on the performance of the DNN. We perform explicit propagation of uncertainty in input data through all layers of the DNN, as well as implicit prediction of output uncertainty to capture the former. Furthermore, we adopt Monte Carlo dropout to capture uncertainty in DNN parameters. We demonstrate the approach for fault detection of power lines with a physics-based model, two types of input data and three different neural network architectures. We compare the performance of such \textit{uncertainty-aware} probabilistic models with their deterministic counterparts. The results show that the probabilistic models provide important information regarding the confidence of predictions, while also delivering an improvement in performance over deterministic models.
\end{abstract}



\end{frontmatter}


\section{Introduction}
Deep neural networks (DNNs) have been immensely successful for building autonomous systems for forecasting~\cite{gasparin2022deep}, monitoring~\cite{ali2022structural}, fault detection~\cite{li2022perspective} and control \cite{oh2022integration} to name a few. The performance of DNNs is heavily reliant on the fidelity and quantity of data used for training. In applications where collecting vast quantities of data can be prohibitively expensive or time-consuming, a digital twin of the system can be used to generate the data required for training DNNs. Such a digital twin can be a physics-based model or data-driven model that can be used to mimic the behaviour of the system. This approach finds applications in structural health monitoring~\cite{dang2021cloud}, quality prediction~\cite{liu2022digital} and fault diagnosis~\cite{wang2022digital} among others.

The parameters of the physics-based or data-driven digital twin are tuned with finite data collected from the system, and hence have uncertainty associated with them. This can reflect in the twin's behaviour in the form of uncertainty associated with the generated data. A physics-based model that employs reduced-order modeling (e.g., to simplify computationally demanding processes in the underlying system) can also have uncertainty in its predictions~\cite{freitas2020parametric}. As a result, any downstream tasks performed with data generated from a digital twin will be influenced by this uncertainty. In addition, DNNs also exhibit uncertainty in the parameters learnt from data, which can ultimately affect their performance. In fact, many state-of-the-art DNNs have been shown to be overconfident in their predictions~\cite{guo2017calibration,minderer2021revisiting}.

DNN-based models for fault diagnosis and monitoring that are developed with data generated from (physics-based or data-driven) digital twins therefore encounter uncertainty in both the data used for training and the parameters learnt during training. These uncertainties can result in (a) over- or under-confidence of the model in its predictions and/or (b) poor performance of the model on unseen data generated by the underlying real-life system. In such scenarios, it is important to exercise caution in interpreting the model's predictions for downstream decision making. This can be achieved with a framework that accounts for these two sources of uncertainty and estimates their impact on the predictions of the DNN model. This article proposes a method to quantify this impact by explicitly incorporating uncertainty at the time of training as well as evaluation. The proposed approach is built on well-established methods of uncertainty quantification in the deep learning literature, and can be used to build uncertainty-aware monitoring and diagnosis tools with digital twin and DNN-based pipelines. In this article, fault diagnosis of electric power lines with a physics-based model of a real line and three different DNN architectures is used as an application to demonstrate the proposed approach.

\subsection{Related Work}
Uncertainty quantification (UQ) is an important aspect of all model-building exercises and has been emphasised as an evaluation criterion in applications ranging from medicine~\cite{begoli2019need} to autonomous vehicles~\cite{michelmore2018evaluating}. Models built from data are subject to aleatoric uncertainty which stems from randomness in the data generating process, sensing mechanism, etc. This type of uncertainty is inherent to the system and cannot be reduced by collecting more measurements. On the other hand, epistemic uncertainty stems from the inadequacy of a model to explain the underlying system, and can be reduced by using more data to train the model. As a result, UQ of both digital twin models and DNN models has already been studied in the literature. The authors in~\cite{woodcock2020uncertainty} made use of Monte Carlo co-simulations with a digital twin and adopted statistical analysis tools to quantify the uncertainty in controlling an agricultural vehicle. A Bayesian model was used for surrogate modelling and UQ of convolutional neural networks learning to solve stochastic partial differential equations in~\cite{zhu2018bayesian}. The uncertainty of surrogate models in reliability-based design optimisation was quantified in~\cite{li2019surrogate}. UQ of computational thermodynamics models was performed in \cite{otis2022uncertainty}. Bayesian method-based model calibration and UQ was performed for a physics-based model for predicting the evolution of precipitates during a heat treatment in~\cite{tapia2017bayesian}. Physics-based digital twins can additionally have parametric or structural uncertainty in the model as well as aleatoric uncertainty. These uncertainties have been addressed in the literature with a host of methods such as calibration and bias correction~\cite{thelen2022comprehensive}. A physics-constrained deep neural network was developed in~\cite{zhu2019physics} to solve partial differential equations, and uncertainty calibration was included in the optimisation to quantify the output uncertainty.

Uncertainty quantification of DNN models has also attracted significant attention. The authors in~\cite{gast2018lightweight} proposed a computationally light approach to quantify the aleatoric and epistemic uncertainty for DNNs. The authors proposed the use of assumed density filtering for propagating the input uncertainty through all layers of the model. Another computationally inexpensive approach was proposed in~\cite{kendall2017uncertainties}, wherein the authors presented a novel formulation of regression and classification problems in the presence of uncertainty. The authors in~\cite{xiao2019quantifying} found that UQ and explicit incorporation of uncertainty provides an estimate of the confidence of predictions, as well as increases the performance of models for several natural language processing tasks. The use of radial basis functions for estimating model uncertainty of DNNs was proposed in~\cite{van2020uncertainty}. These theoretical advancements have also been adapted for UQ in several applications. The authors in~\cite{aouichaoui2022uncertainty} quantified the uncertainty in predictions of two graph neural network architectures for predicting critical properties of materials. Bootstrapping, ensembling and dropout were adopted to quantify the uncertainty. A Bayesian graph convolutional network was used in~\cite{ryu2019bayesian} for predicting molecular properties, and it was observed that the aleatoric uncertainty exhibited a much more significant impact on the model performance than epistemic uncertainty.

\subsection{Motivation}
UQ of DNN models built from digital twins has not received much attention in the literature. It is, however, important to note that such models are trained with a digital approximation of the underlying system. Digital twins are developed with finite amount of data collected from the system, and can have uncertainty in their structure and/or parameters. In certain instances, this data can be limited in accurately capturing the dynamics of the system, while in others, data corresponding to different operating states of the system might not be available. This can accentuate the inaccuracies in correctly replicating the system dynamics, causing a mismatch between the data seen by the DNN at the time of training, and the data it encounters at the time of deployment. For instance, the authors in~\cite{gjorgiev2022simulation} developed and validated a physics-based model of a power line in Switzerland with real-life measurements. However, the data corresponding to different types and severities of faults in the line was not available. This is commonplace in complex engineered systems that are designed to be robust to external perturbations and operate within desired operating bounds. In such a scenario, it is not possible to validate the twin for all possible operating conditions, and check if there any mismatches between the twin and the real system.

The mismatches between the training and deployment data can expose the DNN to out-of-distribution (OOD) samples, resulting in incorrect predictions. This motivates the need to quantify the reliability of the DNN model for correct interpretation of its predictions. UQ in these situations provides additional information regarding the confidence of DNN's predictions, which can be used to inform downstream decision making. 
Specifically, a highly confident prediction of the DNN can be interpreted as being reliable, while poor confidence can point to OOD data encountered by the model. This information is useful in two aspects - (1) it can be used to suggest the end-user to exercise caution in downstream decision making, and (2) such OOD data can be recorded to fine-tune the model further and improve its performance in the future. This article caters to this cause and proposes a system-agnostic framework for UQ of DNNs built from digital models of real-life systems.

\subsection{Contributions}
This article makes the following contributions:
\begin{enumerate}
    \item \textbf{Uncertainty quantification:} We propose a generic UQ framework that is agnostic to the underlying data generating process. The overall uncertainty in predictions of a DNN model is decomposed into two parts:
    \begin{enumerate}
        \item \textit{aleatoric uncertainty} is quantified with (i) explicit propagation of uncertainty and (ii) implicit prediction of uncertainty. The former approach can be used when uncertainty in training data is available, while the latter approach can be used in the absence of such information.
        \item \textit{epistemic uncertainty} is quantified with the popular Monte Carlo Dropout approach. 
    \end{enumerate}
    \item \textbf{Case study:} We apply the proposed UQ pipeline for fault diagnosis of power lines with data generated from a digital twin and three types of DNN architectures. Multiple instances of the digital twin are obtained and the information from these instances is used to inform the DNN training process.
    \item \textbf{Analysis and comparison:} We present a detailed analysis of the performance of uncertainty-aware probabilistic models. A comparison of the two types of aleatoric uncertainty quantification methods is performed. The performance of probabilistic models is compared with their deterministic counterparts.
    \item{\textbf{Out-of-distribution analysis:}} We calculate the performance of probabilistic and deterministic models on data generated from different instances of the digital twin. In addition, we evaluate the robustness of the models to out-of-distribution (OOD) samples.
\end{enumerate}

The rest of the article is organised as follows: Section 2 presents the uncertainty quantification pipeline. The case study used to demonstrate the proposed approach is discussed in Section 3, followed by results of the experiments and detailed analyses in Section 4. A discussion of some insights from the experimental studies is presented in Section 5. The article ends with concluding remarks in Section 6.

\section{UQ of digital twin-driven DNN models}
DNN models build for health monitoring and fault diagnosis of systems can be categorised as solving (a) a regression problem (when the model predicts a continuous variable indicating the health of the system, e.g., remaining useful lifetime of a battery) or (b) a classification problem (when the model predicts a discrete variable representing the state of the system, e.g., healthy or faulty ball bearings). In this section, we exploit well-established UQ techniques in the deep learning literature to develop a system-agnostic and problem-agnostic framework for UQ of DNN-based monitoring and diagnosis models that are built from digital twins.

Consider a system $\Xi$ for which we wish to develop a monitoring or diagnosis model. This can be a dynamic system, e.g., a chemical refinery or a static system e.g., a bridge. Let $f(x;\theta)$ represent a DNN-based monitoring/diagnosis model that is parameterised in $\theta$, and takes $x\in\mathcal{X}$ as input to produce $y\in\mathcal{Y}$ as output. Here, $x$ can represent measurements of variables of the system that are used to predict the health/state $y$. In the absence of training data available from the real-life system, the digital twin $\hat{\Xi}$ of the system generates synthetic data: $\tilde{x}\in\tilde{\mathcal{X}}$ and $\tilde{y}\in\tilde{\mathcal{Y}}$. After training the parameters $\theta$ of the model with data from $(\tilde{\mathcal{X}}, \tilde{\mathcal{Y}})$, the model is deployed on the real-life system to make predictions with data from $\mathcal{X}$.

Let us consider that $f(x;\theta)$ consists of $L$ layers such that the activation $a^{l}$ of the $l^{th}$ layer can be expressed as:
\begin{equation}
    a^l=g_l(a^{l-1};\theta_l) \label{eq:al}
\end{equation}
Here, $g_l(\cdot)$ and $\theta_l$ represent the non-linear activation and parameters of the $l^{th}$ layer respectively. The output of the model can be expressed as:
\begin{eqnarray}
y&=&f(a^0; \theta)=a^l \nonumber \\
&=&g_L(a^{L-1}; \theta_L)\\
&=&g_L\left(g_{L-1}\left(\hdots g_1(a^0; \theta_1)\hdots; \theta_{L-1}\right); \theta_L\right) \nonumber
\end{eqnarray}
The total uncertainty in the prediction $y_i=f(x_i,\theta)$ of the model arises from the uncertainty in $x_i$ (aleatoric uncertainty) and $\theta$ (epistemic uncertainty).

\subsection{Aleatoric Uncertainty}
The aleatoric uncertainty in $f(x;\theta)$ arises from the fact that $x$ is a random variable drawn from a distribution $p_x(x)$. Quantifying the impact of this source of uncertainty can be achieved in two approaches, as described next.

\subsubsection{Explicit propagation of input uncertainty}
In this approach, it is assumed that information regarding the uncertainty in inputs $x$, specifically the mean $\mu_x$ and variance $\Sigma_x$ is available beforehand. The objective is to predict the mean and variance of outputs, i.e., $\mu_y$ and $\Sigma_y$ respectively. In order to achieve this, we exploit assumed density filtering (ADF) that can be used to (approximately) propagate the uncertainty through each layer of the model.

Let $a^{0:l}$ represent the activations of the model up to the $l^{th}$ layer. Then the joint probability density function (PDF) of all the activations of the model can be expressed as:
\begin{eqnarray}
    p(a^{0:L})&=&p(a^0, a^1, \hdots, a^L)
    \\ &=&p(a^0)\prod_{l=1}^L p(a^{l}|a^{l-1})\label{eq:p_all}
\end{eqnarray}
This implies that estimating the joint PDF of all activations involves estimating $p(a^l|a^{l-1}), \forall l\in[1,L]$. The objective of ADF is to tractably approximate this PDF, one layer at a time. To that end, let us assume that $x=a^0\sim\mathcal{N}(\mu_0, \Sigma_0)$. Then let us approximate the activations of subsequent layers as:
\begin{equation}
    p(a^l|a^{l-1})\approx q(a^l)=\prod_j\mathcal{N}(a^l_j; \mu_j^l, v_j^l) \label{eq:p}
\end{equation}
where $\mu_j^l$ and $v_j^l$ represent the mean and variance of activation of $j^{th}$ neuron in the $l^{th}$ layer. Note that the activations of individual neurons in a layer conditioned on activations of previous layer are independent of each other, i.e., $a_j^l\perp a_i^l|a^{l-1}$, $\forall i\neq j$. Thus, the only approximation in Equation \eqref{eq:p} is due to the assumption of normality of individual activations. Substituing this into Eq.~\eqref{eq:p_all}, we obtain the approximate joint PDF as:
\begin{equation}
    \tilde{p}(a^{0:L})=p(a^0)\prod_{l=1}^Lq(a^l)
\end{equation}
The best approximate distribution $q^*(a^l)$ can be obtained by minimising the KL-divergence between $\tilde{p}(a^{0:L})$ and $q(a^{0:L})$~\cite{gast2018lightweight}. It has been shown that this requires matching the moments of the two PDFs~\cite{minka2001family}, which under the assumption of Gaussian distributions, reduces to estimating the mean and variance as follows:
\begin{eqnarray}
    \mu_a^l&=&\mathbb{E}\left[g_l(a^{l-1}; \theta_l)\right]
    \\ v_a^l&=& \text{var}\left[g_l(a^{l-1}; \theta_l)\right]
\end{eqnarray}
The above approach allows variational uncertainty propagation, i.e., approximating the PDF of activations of each layer in the model and propagating the approximate PDF through all layers. This approach essentially replaces the deterministic layers in the model by their variational counterparts that accept the mean and variance of the inputs and generate mean and variance of outputs. The variational approximations of common nonlinear activation layers have already been developed and can be found in~\cite{gast2018lightweight} for both regression and classification problems. A more detailed account of ADF for propagation of input uncertainty can be found in \cite{gast2018lightweight}.

This approach requires the user to train the network with $(\mu_x, \sigma_x^2, y)$ as data samples. An estimate of the variance of the inputs can be obtained from noise characteristics, sensor resolution, etc. However, in dynamic systems, the variance can change over time and can vary across different entries of a single sample. In order to address such heteroskedastic signals with time-varying properties, an implicit prediction approach can be useful.

\begin{figure}
    \centering
    \subfigure[Explicit propagation of input uncertainty ($\mu^l$ and $\Sigma^l$ are mean and variance of activations of $l^{th}$ layer respectively)]{
    \includegraphics[width=0.44\textwidth]{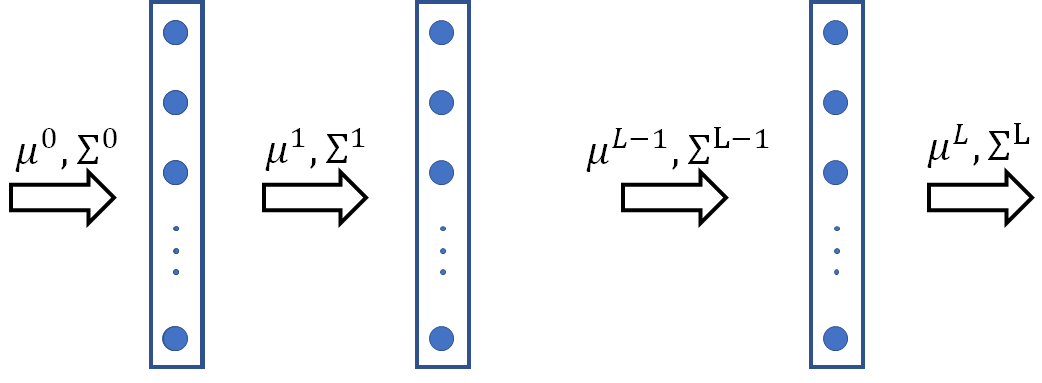}
    }
    \subfigure[Implicit prediction of output uncertainty ($a^l$ is the activation of $l^{th}$ layer; $T$ samples drawn from $\mathcal{N}(\mu^{L-1},\Sigma^{L-1})$ can be used to obtain $\mu^L$ and $\Sigma^L$)]{
    \includegraphics[width=0.44\textwidth]{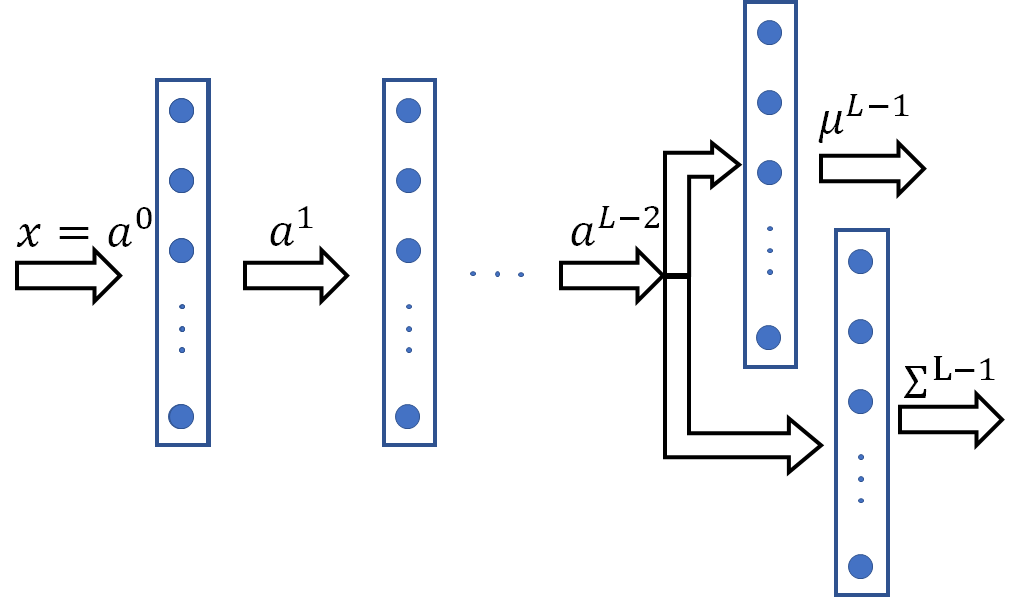}
    }
    \caption{Quantification of aleatoric uncertainty (rectangles with filled circles are DNN layers with neurons; arrows are connections between two layers; the layers can be of any type)}
    \label{fig:aleatoric}
\end{figure}

\subsubsection{Implicit prediction of output uncertainty}
The impact of aleatoric uncertainty on the outputs of the model can also be quantified by implicitly predicting the variance of outputs. In this approach, the dependence of variance on the actual value of $x$, i.e., the heteroskedastic nature of aleatoric uncertainty is considered in the problem formulation. Here, the final layer of the model is redesigned to predict two quantities that are interpreted as mean and variance of the activations. Specifically, as opposed to a deterministic network that predicts $a^L$, the probabilistic network predicts $\mu_a^L$ and $\Sigma_a^L$ in the final layer of the model~\cite{kendall2017uncertainties}. In the case of regression (monitoring models), they represent the mean and variance of predictions of the model. The parameters of the model are then trained by minimising the negative likelihood (NLL), which under the assumption of Gaussian distribution can be expressed as follow~\cite{kendall2017uncertainties}:
\begin{equation}
    \mathcal{L}_{reg}=\frac{1}{n_y}\sum_{i=1}^{n_y}\frac{1}{2\left(\sigma_{a,i}^L\right)^2}||y_i-\mu_{a_i^L}||^2+\frac{1}{2}\log\left(\sigma_{a,i}^L\right)^2
\end{equation}
Here, $\mu_{a,i}^L$ and $\left(\sigma_{a,i}^L\right)^2$ represent the predicted mean and variance of activation for the $i^{th}$ node of the final layer respectively, and $n_y$ is the dimension of $y$.

On the other hand, for classification problem (fault diagnosis models), the activations $a^L$ represent the logits of the model that are passed through the softmax function to obtain the class probabilities for the input. In such a scenario, $\mu_a^L$ and $\Sigma_a^L$ represent the mean and variance of logits, with an assumption of Gaussian distribution, one can draw $T$ samples of logits from $\mathcal{N}(\mu_a^L, \Sigma_a^L)$ as follows:
\begin{equation}
    \tilde{a}_t^L=\mu_a^L+\sigma_a^L\cdot\epsilon_t \label{eq:sampling}
\end{equation}
Here, $\tilde{a}_t^L$ represents a sample logit for $t={1,2,3,\hdots,T}$ and $\epsilon_t\sim\mathcal{N}(\mathbf{0}, I_{n_y})$. Then, the parameters of the model can be trained to minimise the NLL with the following loss function \cite{kendall2017uncertainties}:
\begin{equation}
    \mathcal{L}_{cls}=\frac{1}{T}\sum_{t=1}^T\exp\left(\tilde{a}_{t, c}^L - \log\sum_{i=1}^{n_y}\exp(\tilde{a}_{t, i}^L)\right)
\end{equation}
where $\tilde{a}_{t,i}^L$ represents the $i^{th}$ component of $\tilde{a}_t^L$ and $c$ represents the true class of the input data $x$. This approach was proposed for predicting output uncertainty in computer vision applications in~\cite{kendall2017uncertainties}, and has been shown to be a learned loss attenuation formulation of regression and classification problems in the presence of uncertainty. As opposed to ADF, this allows training and deployment of the model without providing any additional information on uncertainty in the inputs.

\subsection{Epistemic Uncertainty}
The epistemic uncertainty in $f(x;\theta)$ arises from the fact that the parameters $\theta$ are learnt from limited training data and are hence probabilistic. Monte Carlo dropout developed in~\cite{gal2016dropout} is a widely adopted approach to quantify the epistemic uncertainty of DNNs. This approach performs $K$ forward passes of a test sample $x$ with a random dropout of neurons in the model, and quantifies epistemic uncertainty by estimating the variance of predictions across all $K$ forward passes. This can be achieved as follows:
\begin{subequations}
    \begin{eqnarray}
        y_k&=&f(x;\theta^k), k=1,2,3,\hdots,K \label{eq:ep_1}\\
        \mu_y&=&\frac{1}{K}\sum_{k=1}^Ky_k \\
        \sigma_{ep}^2&=&\frac{1}{K-1}\sum_{k=1}^K(y_k-\mu_y)^2 \label{eq:ep_3}
    \end{eqnarray}
\end{subequations}
Here, $\theta^k$ is obtained by randomly masking neurons in the $k^{th}$ forward pass through the model and $\sigma_{ep}^2$ is the variance in predictions because of epistemic uncertainty.

\subsection{Total Uncertainty}
The calculation of total uncertainty involves summation of the aleatoric and epistemic components. However, depending on the type of problem solved (regression or classification), this calculation can be slightly different. In case of regression, the aleatoric variance can be obtained from explicit propagation or implicit prediction as follows:
\begin{align}
    \sigma_{y|k}^2=v_{a|k}^L, && \text{or} && \sigma_{y|k}^2=\left(\sigma_{a|k}^L\right)^2
\end{align}
where all quantities are calculated for the $k^{th}$ forward pass. Then, the average aleatoric variance can be obtained as:
\begin{equation}
    \sigma_{al}^2=\frac{1}{K}\sum_{k=1}^K\sigma_{y|k}^2
\end{equation}
In case of a classification model, we can obtain the aleatoric variance through the explicit and implicit methods as:
\begin{align}
    \sigma_{y|k}^2=v_{a|k}^L, && \text{or} && \sigma_{y|k}^2=\text{var}(\text{softmax}(\tilde{a}_k^L))
\end{align}
Note here that in the implicit case, every forward pass involves drawing $T$ samples according to Eq.~\eqref{eq:sampling}.

The epistemic variance for both models can be obtained from Eqs.~\eqref{eq:ep_1}-\eqref{eq:ep_3}. Finlay, the total variance can then be obtained as:
\begin{equation}
    \sigma_{total}^2=\sigma_{al}^2+\sigma_{ep}^2
    \label{eq:total_variance}
\end{equation}
The framework proposed above is independent of the underlying system, and can handle both regression and classification problems. In the next section, we use a case study of fault diagnosis of a power line to demonstrate the applicability of the above approach.

\section{Application to fault diagnosis}
\subsection{System description}
The physical system considered as a case study in this article is a 220 KV power line in Switzerland that stretches over $26.53$ km and is made of $26$ segments, each segment upgraded at a different time, and hence of different age. Measurements of current at the two ends of the power line are used to build a physics-based model of the system in MATLAB/Simulink. The model is calibrated to accurately depict the leakage current measurements using genetic algorithm (GA). The line is known to be operating under normal conditions (healthy system) at the time of data collection. Faults are synthetically introduced into the physics-based model by scaling the capacitance of the line segments with a factor ranging from $2$ to $10$. In addition, healthy data is also obtained from simulations with the scaling factor $<2$. Faults are further introduced in each of the $26$ segments, resulting in a total of $27$ classes (faulty and healthy). The interested reader is referred to \cite{gjorgiev2022simulation} for a more detailed discussion of the system, the physics-based model and the synthetic data generation process.

In order to account for the uncertainty in the digital twin -- in this case, the physics-based model -- we build $10$ instances of the twin with GA. In other words, we calibrate the physics-based model of the line multiple times. Since the GA is a stochastic approach, we obtained 10 different sets of tuning parameters, which are then used to create 10 instances of the physics-based model. Each model instance is then used to generate $15000$ samples of leakage current signal spanning $0.2$ seconds, i.e., $10$ cycles. These samples represent either healthy or faulty insulation in one of the $26$ segments of the line.

\subsection{DNN models}
This article considers three different DNN architectures, i.e., fully connected (FC), 1-dimensional convolutional (Conv1d) and 2-dimensional convolutional (Conv2d). All models are built in a top-down manner by starting with only one hidden layer and incrementally adding layers till a desirable performance is achieved. This is performed to be as parsimonious as possible in building the models. The FC model has one hidden layer with $50$ neurons. The Conv1d model has 4 convolutional layers with $4$, $2$, $2$ and $2$ filters followed by two fully connected layers with $100$ and $27$ neurons respectively. The kernel size of all convolutional layers in this model is set to $15$. The Conv2d model has $3$ convolutional layers with $16$, $8$ and $4$ filters respectively of kernel size $(3, 3)$ followed by two fully connected layers with $100$ and $27$ neurons respectively. A stride of $1$ is used for Conv1d models, while a stride of $2$ is used for Conv2d models. Batch normalisation and dropout with probability of $0.2$ are employed after all layers in the model except the prediction layers to improve convergence and prevent overfitting, respectively. All layers are constructed with the rectified linear unit activation. The FC and Conv1d models take the sinusoidal output (current signal) of the twin as input, while the Conv2d model takes the spectrogram of the current as input.

For each architecture, we train (a) a deterministic model, (b) an uncertainty-aware model with explicit propagation of input uncertainty and (c) an uncertainty-aware model with implicit prediction of output uncertainty. In the following discussion, we refer to these models as plain, ADF (assumed density filtering) and HET (heteroskedastic) models for the sake of clarity. The plain and ADF models with FC, Conv1d and Conv2d architectures have $81527$, $20547$ and $15267$ trainable parameters respectively. On the other hand, since the last layer of HET models predict the mean and variance of activations, these models have $82904$, $23274$ and $17994$ trainable parameters for the FC, Conv1d and Conv2d architectures respectively.

\subsection{Training and Evaluation}
All experiments are performed with PyTorch on a workstation with NVIDIA RTX 3070 GPU and 128 GB of RAM. The training of all models is performed with Adam optimiser with a learning rate of $0.001$ for $500$ epochs and a batch size of $128$ samples. A learning rate scheduler that reduces the learning rate when the loss does not decrease for $50$ epochs is also employed. A training-validation-testing split of $0.7$-$0.1$-$0.2$ is used in all experiments. We perform two types of training to evaluate and compare the performance of the resulting models. First, we use the data generate by any one instance of the twin $\hat{\Xi}_i$, $i\in[1, 10]$ and train the machine learning (ML) model. The performance of this ML model on the test subset of the dataset as well as the data from all other twins $\hat{\Xi}_j$, $j\neq i$ is calculated. In the second type of training, a data from a subset of twins is used to train the ML model, and its performance is evaluated on the test subset as well as data from other twins.

In order to evaluate and compare different models, we perform four types of analyses. First, we compare the classification accuracy of the different models and highlight the architectures and training methods that deliver the best performance for the system. Second, we evaluate the calibration of the models with reliability diagrams. This allows one to evaluate whether the outputs of the classification models can be safely interpreted as class probabilities. Third, we perform comparative analysis of the aleatoric and epistemic uncertainty of the different models. The results reveal the relative contribution of these sources to the total uncertainty, and provide additional information on prediction confidence. Finally, the generalisability of the models built from synthetic data to different instances of the digital twin is studied.

\section{Results}
In this section, we present the results of our analyses and highlight the main findings.

\subsection{Performance of uncertainty-aware and deterministic models}
The accuracy of the models are summarised in Table~\ref{tab:acc}. The plain and HET models are trained and tested with data generated from one digital twin, specifically $\hat{\Xi}_1$. On the other hand, the ADF models are trained and tested with mean and variance calculated from $10$ digital twins, i.e., $\hat{\Xi}_i,\forall i\in[1,10]$. Thus, the main difference is that for training plain and HET models, we only use the data of a single instance, while for training the ADF models, we utilise the data from all instances of the twin. It can be observed from Table~\ref{tab:acc} that for the FC architecture, the deterministic model has the worst accuracy of $79\%$, followed by HET model with $87\%$ and ADF model with $98\%$. The ADF models outperform the Plain and HET models for all architectures. This is an expected outcome, since ADF models use more information for training than Plain and HET models. The HET models are sensitive to the type of architecture used, and have the worst performance for convolutional architectures, while outperforming the plain models for FC architectures. It was also observed that the performance of the HET models is very sensitive to model initialisation, while the Plain and ADF models are highly robust to initialisation. It must be noted here that we compare only the classification accuracy in this section, and do not consider the fact that ADF and HET models provide additional information regarding the confidence of the predictions, which is discussed in Section~\ref{sec:uncertainty}.

We also observe that although the FC and Conv1d architectures both take raw time series data as input, the latter performs considerably better for the Plain model. This can be attributed to the fact that while the FC architecture looks at relationships between all variables and can be overwhelmed with such relations, the Conv1d architecture extracts patterns relevant to temporal evolution of the data and can identify important information for classification. The Conv2d architecture, on the other hand, uses spectrograms as input, and receives data that has already been processed to extract certain features, i.e., the temporal evolution of spectral content of the signal. As a result, this architecture performs better than the FC and Conv1d architectures for Plain and ADF models.

\begin{table}
    \centering
    \caption{Accuracy of uncertainty-aware and deterministic models}
    \begin{tabular}{cccc}
        \hline
        Architecture & Plain & ADF & HET \\
        \hline
        \hline
        FC & 0.79 & 0.98 & 0.87 \\
        Conv1d & 0.98 & 0.99 & 0.45 \\
        Conv2d & 0.85 & 0.99 & 0.72 \\
        \hline
    \end{tabular}
    \label{tab:acc}
\end{table}

\subsection{Reliability of uncertainty-aware and deterministic models}
\begin{figure*}
    \centering
    \subfigure[FC models]{
        \includegraphics[width=0.3\textwidth]{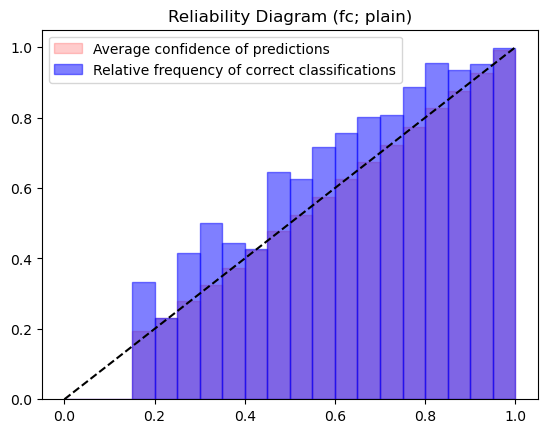}
        \includegraphics[width=0.3\textwidth]{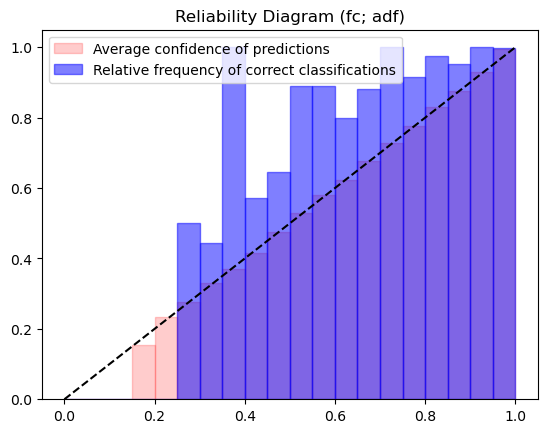}
        \includegraphics[width=0.3\textwidth]{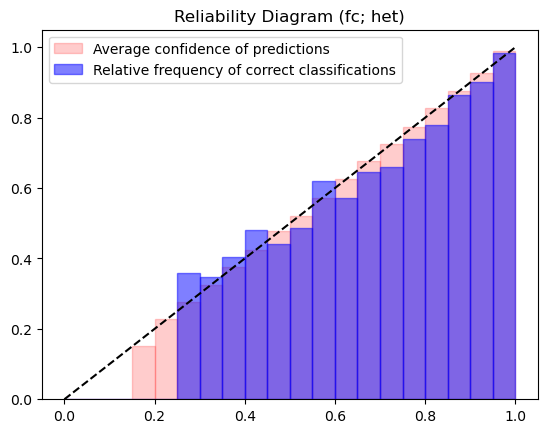}
    }
    \subfigure[Conv1d models]{
        \includegraphics[width=0.3\textwidth]{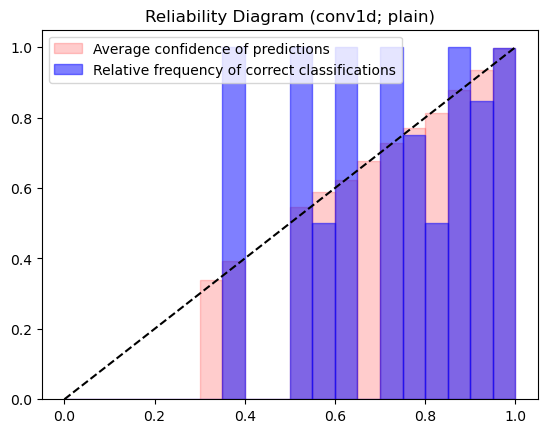}
        \includegraphics[width=0.3\textwidth]{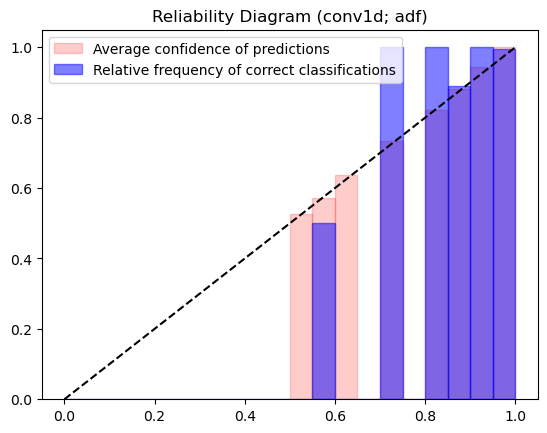}
        \includegraphics[width=0.3\textwidth]{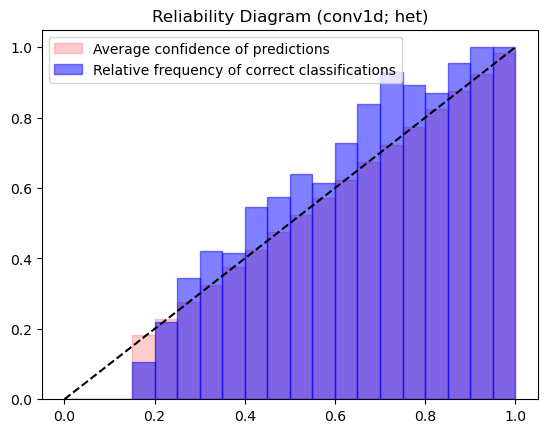}
    }
    \subfigure[Conv2d models]{
        \includegraphics[width=0.3\textwidth]{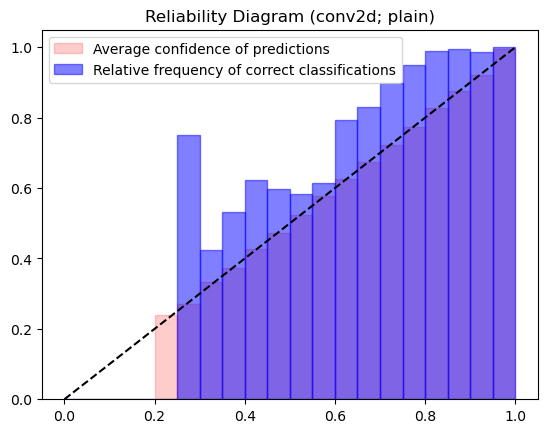}
        \includegraphics[width=0.3\textwidth]{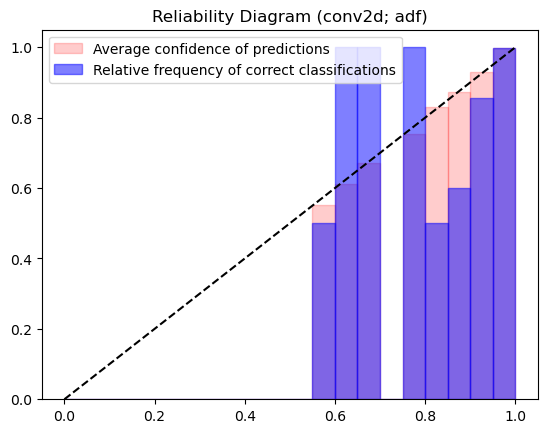}
        \includegraphics[width=0.3\textwidth]{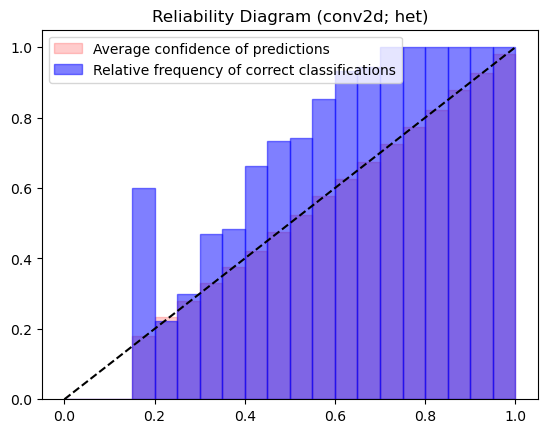}
    }
    \caption{Reliability diagrams of plain, ADF and HET models}
    \label{fig:reliability_diagram}
\end{figure*}

The reliability diagrams of all the models are presented in Fig.~\ref{fig:reliability_diagram}. The reliability diagram compares the predicted probability of a class (pink bars) with the relative frequency of correct classification (blue bars) and is referred to as calibration of the ML model. A perfectly calibrated model is expected to exhibit a reliability diagram where both the quantities are equal, suggesting that the predictions from the model can reliably be interpreted as class probabilities. A poorly calibrated model, on the other hand could have confidence greater or less than the accuracy, resulting in over-confident and under-confident models respectively.

Figure~\ref{fig:reliability_diagram} shows that for the FC architecture, the Plain and HET models are relatively well calibrated compared to the ADF model. We also observe that in most of the cases, the ADF model has a lower predicted probability than the frequency of correct classifications, suggesting that it is mostly under-confident. The Conv1d architecture exhibits a poor calibration for Plain and ADF models that are also under-confident for most cases and over-confident in some. This is a useful observation for the case study, since it implies that in most of the cases, the model is not overly confident in its predictions and thus, can be relied upon to make downstream decisions. For the Conv2d architecture, we observe relatively better calibration than the Conv1d counterparts for the Plain and ADF models. The observed miscalibration in the models can be corrected to some extent by employing techniques such as temperature scaling~\cite{guo2017calibration} to re-calibrate the models. However, this is beyond the scope of this article and can be explored in future studies to enhance the reliability of the models.

\subsection{Uncertainty in model predictions} \label{sec:uncertainty}
\begin{figure}
    \centering
    \includegraphics[width=0.44\textwidth]{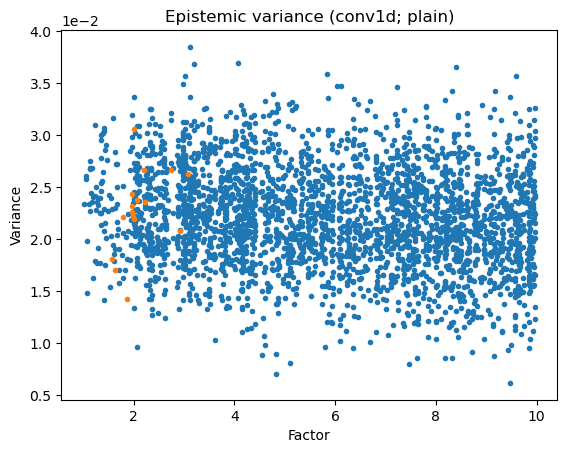}
    \caption{Epistemic variance of Plain Conv1d model (blue points are correctly classified samples; orange points are incorrectly classified samples)}
    \label{fig:epistemic}
\end{figure}

\begin{figure*}
    \centering
    \subfigure[FC models]{
        \includegraphics[width=0.31\textwidth]{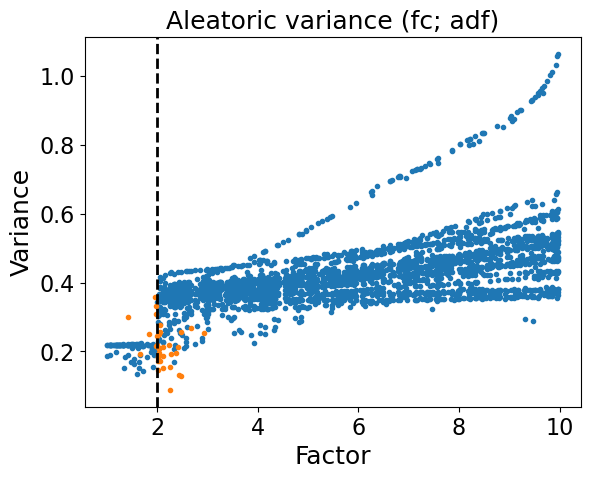}
    }
    \subfigure[Conv1d models]{
        \includegraphics[width=0.31\textwidth]{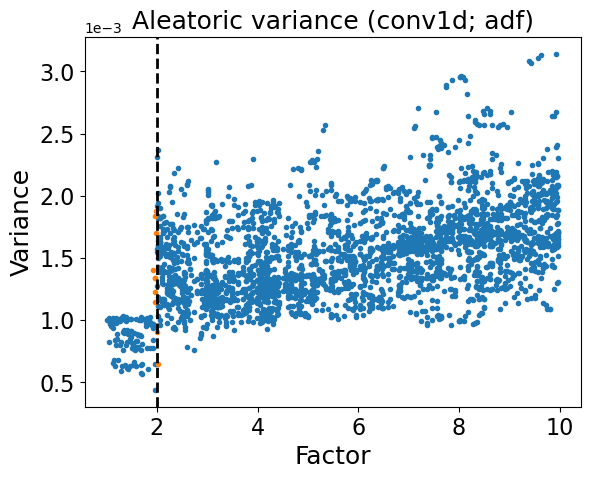}
    }
    \subfigure[Conv2d models]{
        \includegraphics[width=0.3\textwidth]{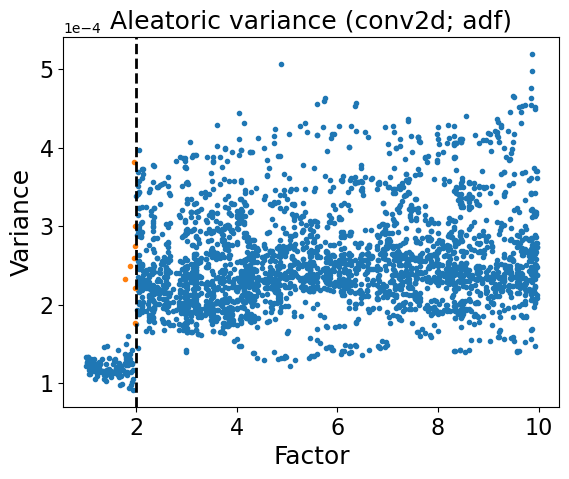}
    }
    \caption{Aleatoric uncertainty (variance) of ADF models (blue points are correctly classified samples; orange points are incorrectly classified samples; the dashed black line separates healthy samples from faulty ones)}
    \label{fig:vars}
\end{figure*}

We observed that the epistemic variance of the deterministic models for all architectures did not exhibit any distinct patterns, and were uniformly distributed in a specific range. This can be observed in Fig.~\ref{fig:epistemic} which shows the epistemic variance for the Plain model with Conv1d architecture as a illustrative case. A similar observation was also made for the aleatoric variance of the HET models (not shown in a figure). The aleatoric variance of the ADF models exhibited interesting patterns, which are shown in Fig.~\ref{fig:vars} shows aleatoric variance of all of the ADF models. Here, the variances are plotted as a function of the scaling factors of capacitance in the digital twin that are used to simulate different healthy and faulty conditions. First, we observe that the variance of the FC model is more than that of the Conv1d model, which is larger than that of the Conv2d model. We also observe that all models exhibit a sudden increase in variance as the scaling factor increases beyond $2$, representing faulty data. This suggests that the model is much more confident in the classification of data for the healthy system than for the faulty system. It is noteworthy that the physics-based models are tuned with a healthy system, and hence are expected to exhibit better fidelity in this region compared to the faulty region (scaling factor $>2$). Although this information is not provided at the time of training, the ADF model is interestingly able to discover the effect of this difference in the data on its own, and thus reflect it in its predicted variances.

Furthermore, we observe that for the faulty region, the variance of predictions increases with the scaling factor. The model thus becomes less confident as the scaling factor increases. This behaviour is true for the FC and Conv1d models, while the Conv2d model appears to be insensitive to the scaling factor. The ADF models are therefore able to provide the most amount of information regarding their predictions, and are also sensitive to the digital twin tuning and data generation processes. We next look at the generalisability of the models to data generated from different digital twins.

\subsection{Generalisability of uncertainty-aware and deterministic models}
The Plain models trained with data from $\hat{\Xi}_1$, when evaluated on the data generated from any other twin, i.e., $\{\hat{\Xi}\}_{i\neq1}$ result in a classification accuracy of $0.03$ for all architectures. This is equivalent to performing a random guess with $27$ classes and indicates no transferrability of the features learnt by the models to different datasets. It must be noted that all models considered here are small networks with less than $100,000$ parameters, and have dropout layers after all layers except the last prediction layers. As a result, this poor generalisability cannot be attributed to over-fitting by the model, and is a direct consequence of the variability in the data generated by different twins. For this case study, training deterministic models on data generated by digital twins is therefore as good as making an naive guess and cannot be relied upon to assess the state of health of the real-life system. The HET models also exhibit a similar behaviour with a classification accuracy of $0.04$ for different datasets.

The ADF models, on the other hand, exhibit better generalisability across different datasets. We trained ADF models with mean and variance calculated from $7$ datasets randomly chosen from the $10$ available datasets and recorded the performance on the test subset. We then tested this model on the mean and variance calculated with the remaining $3$ datasets. We observed that the absolute performance of the ADF models trained with $7$ datasets is the same as that trained with $10$ datasets, and are comparable with that reported in Table~\ref{tab:acc}. Furthermore, difference in their performance on the remaining $3$ datasets was found to be in the range $[0.05, 0.15]$, still providing more than $85\%$ classification accuracy. This decline in performance was further observed to be sensitive to different initialisations and to the type of architecture used. This highlights the value of providing additional information on input data variance at the time of training and using ADF to train the models. However, real-life implementation of ADF models requires that the variance of inputs be provided at the time of deployment, which might not be available in all applications. We provide further discussion along these lines in the following section.

\section{Discussion}
In this section, we discuss some key properties of the probabilistic models and the value of using multiple instances of digital twins.



\textit{Information fusion from multiple twins:}
In the previous section, we train the Plain and HET models with data generation from one digital twin, specifically $\hat{\Xi}_1$, as opposed to the ADF models that are trained with data processed from all digital twins, i.e., $\{\hat{\Xi}_i\}_{i=1}^{10}$. The ADF models therefore see the average (and variance) of the the data generated from $\hat{\Xi}_i$, and it is interesting to investigate if this information fusion from multiple digital twins can also improve the performance of the Plain and HET models. We train Plain and HET models with the mean of data from all digital twins and observe that the models perform comparable to the ADF model for all architectures. This suggests that it is indeed valuable to fuse information from multiple instances of the digital twins to train DNN models to achieve better performance. However, it is also important to provide similarly averaged data at the time of deployment of the model.

\textit{Deployment of ADF models in real life:}
We observe that ADF models provide the best performance as well as insights into the relationship between variance and the data generation process. However, they also require more data to calculate the mean and variance that are used as inputs to the model. While this is possible with digital twins as shown in the previous section, obtaining this information from the real-life system is important to use ADF models in practice. One can utilise measurements over a longer time, divide the data into multiple sets, and estimate the mean and variance from the sets. Furthermore, it is possible to employ sensor noise characteristics to obtain an estimate of the variance in measurements. This is commonly adopted in computer vision applications where the camera resolution can be used to estimate the uncertainty in the pixel values of images. 

We tested the performance of ADF models trained with a fixed value of variance $=0.53$, which is the average value of the variance across all samples. We observe that the performance of ADF models drops only marginally (by $\sim2\%$). We also observe that the dependence of the prediction variance on scaling factors used to simulate faulty data (as discussed in Section~\ref{sec:uncertainty}) also diminishes for the faulty data. On the other hand, the difference between prediction variances for healthy and faulty data is still observed. The models thus lose both performance and some information related to the variance of predictions. Therefore, ensuring that these models are ready for deployment in real-life systems still remains an open question.



\section{Conclusion}
In this article, we identify the challenge of training reliable models for monitoring of systems with data generated from digital twins. We identify that the uncertainty associated with data generated from digital twins can influence the training of a DNN-based monitoring model and affect its reliability at the time of deployment. We propose an uncertainty quantification framework that explicitly accounts for aleatoric and epistemic uncertainty in training DNN models with data generated from digital twins. Specifically, (1) assumed density filtering and heteroscedastic modeling are used to quantify aleatoric uncertainty, and (2) Monte-Carlo dropout is used to capture epistemic uncertainty. We compare and contrast three neural network architectures and two approaches of uncertainty quantification with a case study on fault detection of faulty insulation across power lines. The results show that training a deterministic model with data generated from a digital twin can result in a model that does not generalise well, and might even be comparable to a random guess. Assumed density filtering provides the best models in terms of classification accuracy and generalisability. They also offer insights into the dependence of prediction variance on the conditions used to simulate different healthy and faulty conditions in the digital twin. On the other hand, they require more information at the time of training, which can be obtained from more observations from the system. Heteroscedastic models are found to perform poorly on our case study. However, they provide an elegant approach to predict aleatoric uncertainty, and their applicability in other applications is worth investigating. Future work can investigate alternate uncertainty quantification approaches and loss function formulations to address the issues of performance and generalisability of uncertainty-aware models.

\section*{Funding Information}
This work was supported by the Swiss Federal Office of Energy: “IMAGE - Intelligent Maintenance of Transmission Grid Assets” (Project Nr. SI/502073-01).

\bibliographystyle{elsarticle-num} 
\bibliography{draft_1}

\end{document}